\documentclass[conference]{IEEEtran}
\IEEEoverridecommandlockouts
\usepackage{cite}
\usepackage{amsmath,amssymb,amsfonts}
\usepackage{algorithmic}
\usepackage{graphicx}
\usepackage{textcomp}
\usepackage{xcolor}
\usepackage{tikz}
\usetikzlibrary{arrows.meta,positioning}

\def\BibTeX{{\rm B\kern-.05em{\sc i\kern-.025em b}\kern-.08em
    T\kern-.1667em\lower.7ex\hbox{E}\kern-.125emX}}
\begin{document}

\title{EEG-SpikeAgent: Agentic Closed-Loop Program Synthesis for Automated EEG Spike Detection}

% Put this before \author if you want a short marker command
\newcommand{\affilmark}[1]{\textsuperscript{#1}}

\author{%
\IEEEauthorblockN{%
Sonali Santhosh\affilmark{1},
Kelly Shuhong Yu\affilmark{2},
Eugene Chang\affilmark{2}\\
Jonathan Kim\affilmark{3},
Kie Shidara\affilmark{4},
Danilo Bernardo\affilmark{4,5}
}
\IEEEauthorblockA{\affilmark{1}Dept. of Brain and Cognitive Sciences, Massachusetts Institute of Technology, Cambridge, MA, USA}
\IEEEauthorblockA{\affilmark{2}Dept. of Neuroscience, University of California, Berkeley, Berkeley, CA, USA}
\IEEEauthorblockA{\affilmark{3}Dept. of Neurology and Neurologic Sciences, Stanford University, Palo Alto, CA, USA}
\IEEEauthorblockA{\affilmark{4}Weill Institute of Neurology and Neurosciences, University of California, San Francisco, San Francisco, CA, USA}
\IEEEauthorblockA{\affilmark{5}\texttt{dbernardoj@gmail.com}}
}
% \author{%
% \IEEEauthorblockN{Anonymous Authors}
% \IEEEauthorblockA{Affiliations withheld for double-blind review}
% }
\maketitle

\begin{abstract}
Automated detection of interictal epileptiform discharges in scalp electroencephalography (EEG) is clinically important, but recent high-performing deep-learning models often trade interpretability for accuracy. We introduce EEG-SpikeAgent, a closed-loop program-synthesis framework that uses a large language model (LLM) agentic system to generate signal-processing features for spike detection in scalp EEG. The system iteratively proposes one deterministic EEG feature module at a time, executes the resulting code on EEG to generate tabular features, evaluates performance via a tabular classifier, summarizes run-level metrics, and feeds structured diagnostics back to the model for refinement. Across iterations, EEG-SpikeAgent proposes and refines candidate signal features and decision rules informed by model performance. We evaluated EEG-SpikeAgent on VEPISET, a public 29-channel dataset of 4-second epochs containing 2,516 discharge-containing and 22,933 non-discharge epochs. Across five-fold cross-validation with a gradient-boosted tree classifier, agent-generated features achieved an area under the receiver operating characteristic curve of 0.935, balanced accuracy of 0.699, F1 score of 0.557, sensitivity of 0.401, and specificity of 0.996 at the default operating point. At an operating point with sensitivity 0.80, mean precision was 0.470 and mean specificity was 0.900. Artifact-aware feature generation improved balanced accuracy and F1 score over spike-only feature search. These results indicate that LLM-based program synthesis can automate EEG feature engineering in auditable and inspectable code-driven manner for clinical and methodological review.
\end{abstract}


\section{Introduction}
Automated detection of interictal epileptiform discharges (IEDs) in scalp EEG remains challenging \cite{Janmohamed2022,Diniz2024Meta,Nhu2022Review}. Human interpreters rely on spatial cues such as phase reversals in bipolar chains \cite{Jadeja2021,Kutluay2019Montages} and temporal features such as distinctive spike morphology \cite{ACNSNomenclature2021,Emmady2025,Li2020,MedscapeAbsence2024,StatPearlsIGE2024}. Purely data-driven methods can perform well but often require large labeled corpora, may learn brittle shortcuts, and lack interpretability \cite{Janmohamed2022,Diniz2024Meta,Nhu2022Review}. Physics-based pipelines are interpretable and data-efficient yet typically evolve via slow, manual iteration \cite{Michel2019ESI,Scherg1990,Hamalainen1993}. We previously demonstrated the potential of LLM agents in EEG analysis~\cite{Kim2024Eeg}, and here, we were motivated to use LLM agents to automate manual engineering of physics-based pipeline for EEG analysis.

Recent deep learning systems have substantially improved upon automating EEG marker detection, with several models having achieved expert-level accuracy in identifying interictal epileptiform discharges (IEDs). Nonetheless, these systems continuously raise interpretability and generalizability concerns, as their performance effectiveness may differ based on varying datasets or recording conditions \cite{Li2025Expert,Tjepkema2025Expert,Wong2025Channel}. Feature-engineered EEG pipelines provide a transparent alternative by incorporating predefined characteristics, namely signal morphology, spectral content, spatial organization, and artifact signatures. However, these pipelines often encounter limitations stemming from their dependence on manually curated feature sets and the prolonged refinement processes required from experts \cite{AbdiSargezeh2024Review,Mallick2024Novel,Tuncer2025Explainable}. Together, these tradeoffs raise a need for systems capable of autonomously navigating the EEG feature space while simultaneously preserving interpretable definitions of features and categories relevant to clinical practice.

Here, we develop EEG-SpikeAgent, a program-synthesis approach for automated spike detection. Rather than pre-specifying a detector, we pose EEG spike detection as a \emph{design-search problem} and use a large language model (LLM) in a closed loop to iteratively propose EEG feature detection code, auditable feature code edits, and threshold updates based on structured diagnostics \cite{Madaan2023SelfRefine,Shinn2023Reflexion}. The loop couples automated feature space exploration with an auditable harness ensuring that each generated feature can be inspected, ablated, and/or traced to model behavior.

\section{Methods}
\label{sec:methods}

\begin{figure*}[t]
\centering
\resizebox{\textwidth}{!}{%
\begin{tikzpicture}[
    >=Stealth,
    node distance=9mm and 7mm,
    every node/.style={font=\footnotesize},
    block/.style={draw, rounded corners=2pt, thick,
        align=center, minimum width=30mm, minimum height=11mm,
        inner sep=3pt},
    ctx/.style={block, fill=gray!8},
    llm/.style={block, fill=gray!18},
    host/.style={block, fill=gray!5},
    eval/.style={block, fill=gray!13},
    arrow/.style={->, thick}
]

% Main agentic loop
\node[ctx] (context) {Prompt context\\
code summary, assumptions,\\
history, manifest};

\node[llm, right=of context] (propose) {LLM proposal\\
one complementary,\\
interpretable feature};

\node[llm, right=of propose] (edit) {Apply code edit\\
minimal \texttt{feature\_*.py}\\
module};

\node[host, below=of edit] (execute) {Host execution\\
register feature, run blocks,\\
tabularize fixed split};

\node[eval, left=of execute] (evaluate) {Portfolio evaluation\\
XGB CV; balanced acc.,\\
F1, AUROC, errors};

\node[ctx, left=of evaluate] (diagnose) {Diagnostic summary\\
metrics, confusion matrix,\\
failure cases};

% Loop arrows
\draw[arrow] (context) -- (propose);
\draw[arrow] (propose) -- (edit);
\draw[arrow] (edit) -- (execute);
\draw[arrow] (execute) -- (evaluate);
\draw[arrow] (evaluate) -- (diagnose);
\draw[arrow] (diagnose) -- (context);

\end{tikzpicture}%
}
\caption{Agentic design-search loop used by EEG-SpikeAgent. The LLM proposes and applies a single auditable feature change; the host registers the feature, tabularizes the fixed EEG split, evaluates the portfolio, and feeds diagnostics back into the next prompt.}
\label{fig:agentic-loop}
\end{figure*}
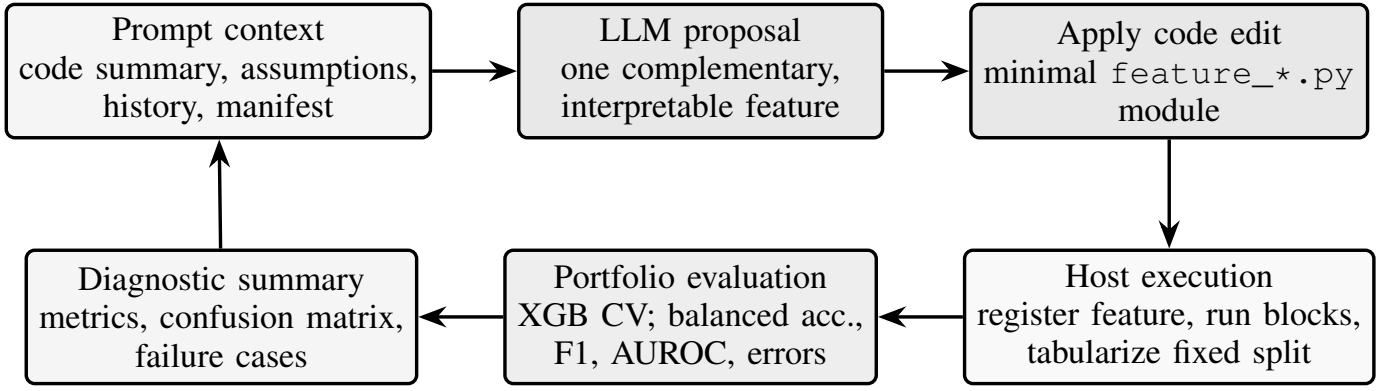

\subsection{Problem formulation and dataset organization}
We cast interictal epileptiform discharge (IED) detection as binary classification on fixed-duration multi-channel EEG epochs $X \in \mathbb{R}^{C \times N}$. We utilized the publicly available VEPISET EEG dataset to validate EEG-SpikeAgent. The VEPISET data are organized into one Non-IED directory and five IED subtype directories (Generalized, Frontal, Temporal, Centro-Parietal, and Occipital IED). For the binary task, Non-IED epochs are labeled ``No Spike'' and all IED subtypes are labeled ``Spike.'' Each example uses a fixed 29-channel order: 19 scalp 10--20 electrodes (Fp1/2, F3/4, C3/4, P3/4, O1/2, F7/8, T3/4, T5/6, Fz, Cz, Pz), bilateral ear/reference channels (PG1, PG2, A1, A2), two ECG channels, and four EMG channels~\cite{Lin2025VEPISET}. Signals are sampled at $f_s=500$~Hz. Epochs shorter than 4.0~s are zero-padded symmetrically along the time axis to 2000 samples; longer epochs are left unchanged.

\subsection{Preprocessing}
The upstream pipeline applies 1--40~Hz Butterworth bandpass filtering, 60~Hz notch filtering, and average re-referencing. Feature executors receive the average-referenced scalp view and the associated 10--20 channel names. Approximate electrode coordinates are available for geometry-aware feature families such as spatial focality, bipolar gradients, propagation, and dipole-topography summaries \cite{Klem1999,Kutluay2019}. Filtering and numerical operations use NumPy/SciPy routines \cite{Virtanen2020}. 

\subsection{LLM-guided design-search protocol}
EEG-SpikeAgent frames feature engineering as an iterative program-synthesis problem \cite{Madaan2023SelfRefine,Shinn2023Reflexion}. Each iteration has two LLM-mediated phases followed by host-side evaluation. First, a proposal prompt supplies the model with the current code summary, dataset assumptions, feature-performance history, and the active feature manifest. The model must propose exactly one complementary feature change. Second, an apply prompt asks a code-editing model to implement that plan as a minimal Python module. Guardrails restrict edits to feature modules named \texttt{~/features/feature\_*.py}; at most one such module may be created or modified per iteration, and core engine, plugin, DSL, and evaluation code are treated as read-only. Artifact-focused feature proposals were scheduled at iterations 5, 10, and 15 to test whether explicit artifact modeling improved downstream IED classification.

Each generated feature module defines a Pydantic configuration model, a registry entry with a unique feature type, and a deterministic executor. The executor maps one EEG epoch to a fixed set of numeric diagnostics. Feature modules are imported automatically, registered into a global feature registry, and tracked in a tabular feature manifest. The manifest fixes the active feature portfolio and column order for subsequent evaluation.

\paragraph{Feature proposal constraints.}
The feature-search prompt asks for tabular features that are sensitive to brief transients and robust to the unknown timing of an IED within a 4~s epoch. The prompt encourages short-window, multiple-instance summaries rather than full-epoch averages, robust per-window or per-channel normalization rather than fixed absolute voltage thresholds, and compact cross-channel aggregations rather than per-sample or per-channel feature explosions. Feature executors do not learn embeddings or fit parameters; learning is deferred to the downstream tabular classifier.

These constraints were chosen to align the search space with the clinical and signal-processing structure of scalp EEG. EEG spikes (or IEDs) are brief, spatially organized events that may occur at any point within a multi-second epoch, so features based only on full-epoch averages risk diluting the transient morphology of interest. Conversely, unconstrained per-sample or high-dimensional channel-specific representations would reduce auditability and increase the risk that the agent would generate brittle or dataset-specific features. By requiring compact, deterministic, and window-aware summaries, EEG-SpikeAgent was encouraged to search for features that preserved clinically meaningful concepts, such as sharpness, focality, field distribution, and artifact coupling, while remaining compatible with standard tabular evaluation. This design also cleanly separated feature discovery from statistical learning. Specifically, the LLM generation phase proposed signal measurements, but classification performance was determined only after those measurements were executed and evaluated by the host pipeline.

\subsection{Tabularization and model evaluation}
After each accepted feature edit, the host process computes or refreshes a per-feature block over the fixed epoch index. Each block stores epoch identifiers, labels, feature names, and the feature matrix for one feature type. For model fitting, blocks are concatenated in manifest order to form a tabular design matrix. This incremental representation allows new features to be evaluated without recomputing unchanged feature blocks.

\begin{figure*}[htbp]
\centering{\includegraphics[width=\textwidth]{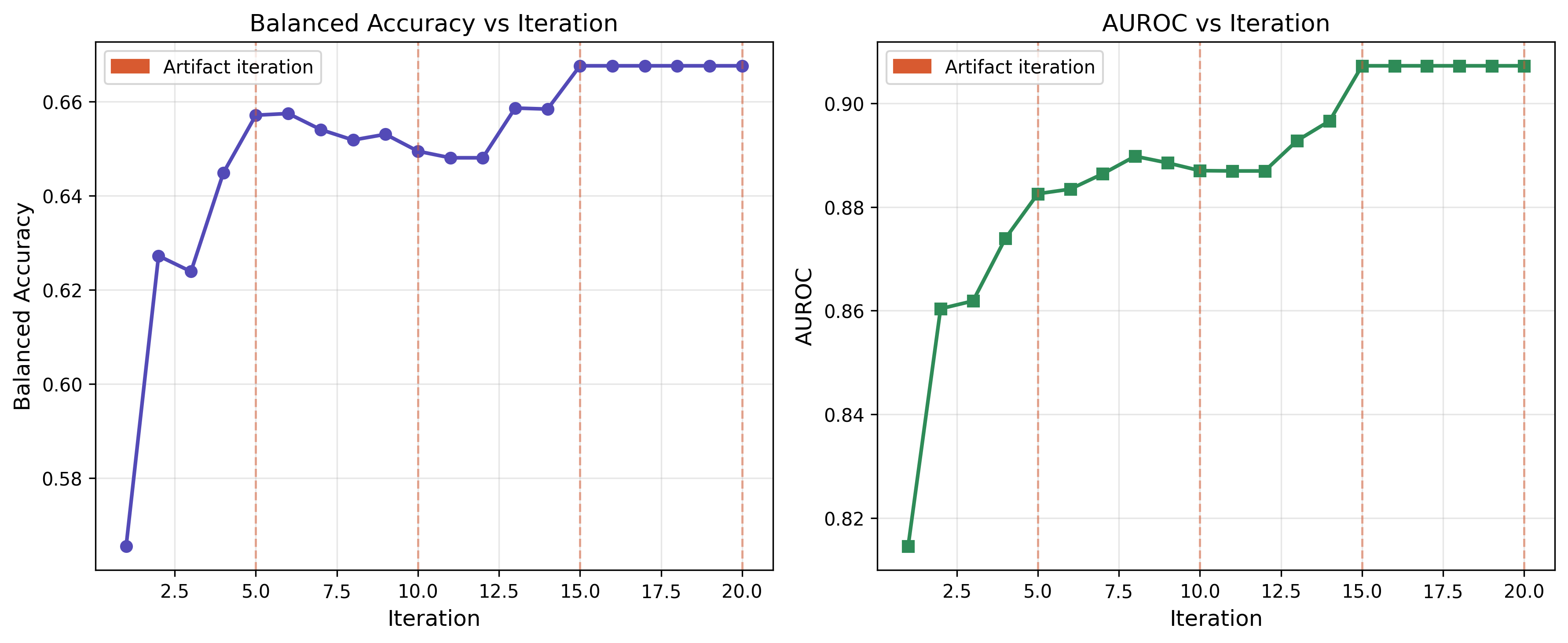}}
\caption{Balanced accuracy trajectory across twenty iterations. Dashed vertical lines indicate artifact-focused feature-generation iterations.}
\label{fig:trajectory}
\end{figure*}

The outer-loop search uses balanced accuracy as the primary optimization signal because the dataset is class-imbalanced. During feature search, candidate portfolios are evaluated with an XGBoost classifier under stratified cross-validation on the training partition, logging balanced accuracy, F1, sensitivity, specificity, AUROC, PRAUC, and the confusion matrix for each iteration. Final reported classifier performance is obtained separately by fitting an XGBoost binary classifier on the training partition per fold and evaluating it on the held-out validation partition \cite{Chen2016XGBoost}. Thus five independent XGBoost classifiers were trained across the 5 CV folds. The XGBoost script uses Spike as the positive class, a binary logistic objective, 300 boosting rounds, maximum tree depth 4, learning rate 0.05, row subsampling 0.8, column subsampling 0.8, histogram tree construction, and random seed 0. Validation probabilities are used to compute ROC and precision--recall curves. Hard predictions are used for accuracy, balanced accuracy, sensitivity, specificity, precision, and F1.

\subsection{Reproducibility}
For each feature proposal the agent records the prompt context, proposal, apply summary, manifest, metrics, and out-of-fold prediction artifacts. Fixed random seeds control dataset splitting and classifier evaluation, to promote reproducibility and traceability of observed artifacts and performance.

\section{Results}
\label{sec:results}

\begin{table*}[htbp]
\caption{Feature Families and Representative Feature Primitives}
\begin{center}
\small
\begin{tabular}{|p{0.28\textwidth}|p{0.64\textwidth}|}
\hline
\textbf{Feature Family} & \textbf{Representative Feature Primitive} \\
\hline

Short-window spike morphology
& Line length; peak amplitude; curvature/sharpness; Teager energy; robust window maxima \\
\hline

Multiscale spike-wave morphology
& Matched-filter responses; multiscale sharpness; spike-to-slow-wave coupling; top-$k$ response summaries \\
\hline

Focal spatial contrast
& Local spatial residuals; nearest-neighbor deviations; focality scores; spread counts \\
\hline

Bipolar phase-reversal evidence
& Bipolar pair differences; polarity opposition; pair sharpness; active pair counts \\
\hline

Regional and laterality summaries
& Left--right asymmetry; temporal/frontal/posterior maxima; temporal-vs-parasagittal ratios; score entropy/Gini \\
\hline

Artifact-reference coupling
& Scalp correlation with ear/ECG/EMG channels; co-activation fractions; reference-envelope products \\
\hline

Channel-quality and burst artifacts
& RMS outliers; peak-to-peak outliers; kurtosis/trend anomalies; burst RMS; inter-channel correlation anomalies \\
\hline

Multiple-instance aggregation
& Max; p95/p99; top-3/top-5 means; event counts; active fractions; longest active run \\
\hline

\end{tabular}
\label{tab:feature_families}
\end{center}
\end{table*}

\subsection{Design Search Outcome}
\label{sec:results-design}
For evaluation of EEG-SpikeAgent we utilized the 29-channel VEPISET IED dataset which contains annotated interictal epileptic data from 84 patients encompassing 28 hours, including 2,516 IED epochs and 22,933 non-IED epochs, each 4 seconds long \cite{Lin2025VEPISET}. We performed 5-fold cross-validation, where within each fold, EEG-SpikeAgent was tasked with creating a new feature set and corresponding XGBoost predictive model. We subsequently refer to a single fold as an independent instantiation. A diagram illustrating the agentic loop is demonstrated in Fig.~\ref{fig:agentic-loop}.  
Without a pre-specified blueprint or direct human intervention, the LLM-guided agentic loop generates between 400 and 500 quantitative EEG features across twenty iterations per independent instantiation. Through the dual-prompting strategy, the agent successfully designated iterations 5, 10, and 15 to artifact-detection features while spike-detection features constituted the remaining iterations. Model performance was observed to increase across iterations and plateau after 15 iterations (Fig.~\ref{fig:trajectory}). Representative features generated are demonstrated in Table ~\ref{tab:feature_families}. The feature registry prevented duplicate feature-type names, and no duplicate modules were accepted into the final manifest per independent instantiation.

\subsection{Detection performance}

\begin{figure}[!t]
\centering{\includegraphics[width=\columnwidth]{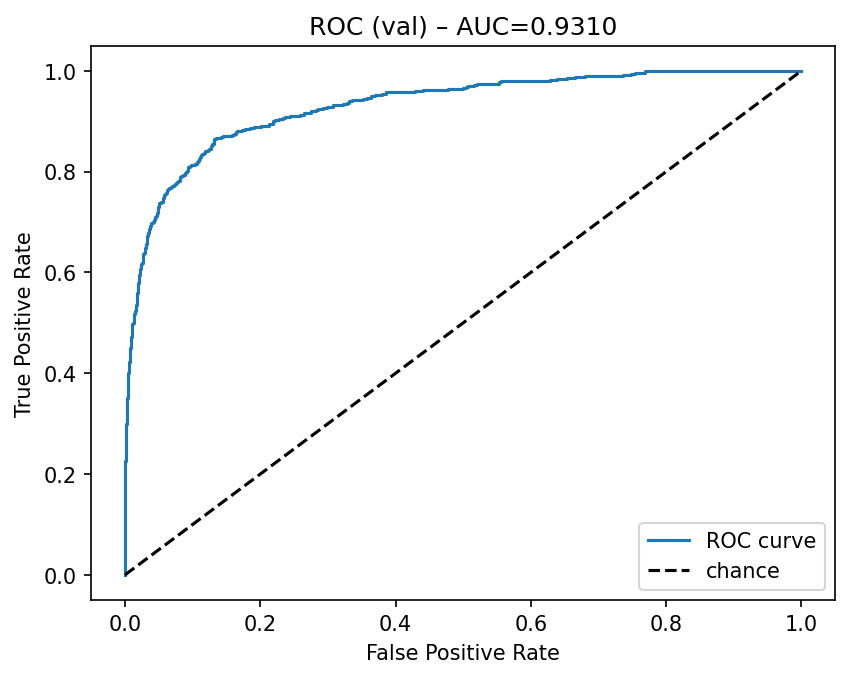}}
\caption{ROC curve (AUROC = 0.93) for the XGBoost classifier.}
\label{fig:auroc}
\end{figure}

\begin{figure}[!t]
\centering{\includegraphics[width=\columnwidth]{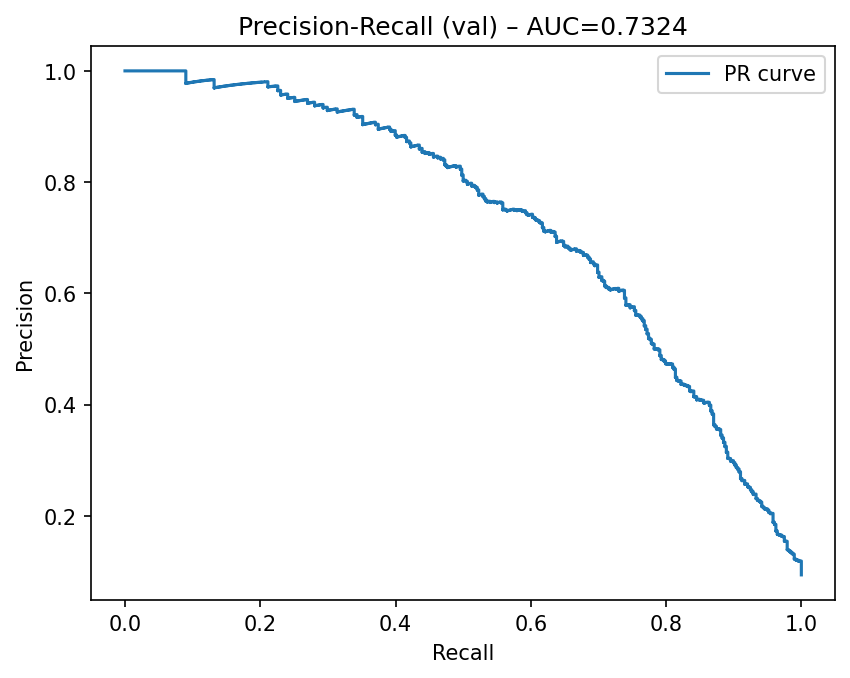}}
\caption{Precision-Recall curve (PRAUC = 0.73) for the XGBoost classifier.}
\label{fig:auprc}
\end{figure}

Across 5-fold cross-validation, the classifier achieved a mean AUROC of 0.935 ± 0.008, with a 95\% CI of 0.925–0.946, indicating strong discrimination between spike and non-spike epochs (Fig.~\ref{fig:auroc}, Fig.~\ref{fig:auprc}). In addition, across folds, the model achieved a balanced accuracy of 0.699 ± 0.016, with a 95\% CI of 0.678–0.719. The classifier attained an F1 score of 0.557 ± 0.034, with a 95\% CI of 0.516–0.599. Sensitivity was 0.401 ± 0.014, with a 95\% CI of 0.361–0.441, while specificity was 0.996 ± 0.001, with a 95\% CI of 0.994–0.997. These results demonstrate that at the default classifier operating point, the model prioritized false-positive suppression over spike sensitivity. At an operating point with sensitivity 0.80, mean precision was 0.470 and mean specificity was 0.900. At the default operating threshold, the classifier favored false-positive suppression over sensitivity. This operating point is consistent with the class imbalance present in VEPISET where non-spike epochs substantially outnumbered spike epochs.

\begin{table}[!t]
\caption{Cross-Validated Model Performance Over Five Folds$^{\mathrm{a}}$}
\label{tab:model_performance}
\centering
\begin{tabular}{lcc}
\hline
\textbf{Metric} & \textbf{Mean $\pm$ SD} & \textbf{95\% CI} \\
\hline
AUROC          & $0.935 \pm 0.008$ & $0.925$--$0.946$ \\
Balanced Acc.  & $0.699 \pm 0.016$ & $0.678$--$0.719$ \\
F1             & $0.557 \pm 0.034$ & $0.516$--$0.599$ \\
Sensitivity    & $0.401 \pm 0.032$ & $0.361$--$0.441$ \\
Specificity    & $0.996 \pm 0.001$ & $0.994$--$0.997$ \\
\hline
\multicolumn{3}{p{0.75\linewidth}}{\footnotesize $^{\mathrm{a}}$ Cross-validated XGBoost model performance over five folds using the default operating point.} \\
\end{tabular}
\end{table}

\subsection{Artifact-aware Ablation} 
Ablation experiments demonstrated that artifact detection features improved balanced accuracy by approximately 1.4 percentage points and F1 Score by 2.3 percentage points compared to the initial spike-only feature detection. However, this improvement was primarily driven by reductions in false-negative errors for true spike events contaminated with artifacts, as depicted in the enhanced sensitivity rates via artifact-aware features. Performance steadied by iteration 16, with the balanced accuracy leveling off at 0.70 and AUROC stabilizing around 0.93 (Fig.~\ref{fig:trajectory}). 

\subsection{Feature Importance Analysis}
To examine which components of the agent-generated feature set contributed most strongly to the final classifier, we ranked all 410 features using gain-based feature importance from the trained XGBoost model. Gain measures the average improvement in the model objective produced by splits that use a given feature, and therefore provides a model-internal summary of which measurements were most useful during tree construction. Because gain-based importance is unsigned and can be biased toward features that offer more split opportunities, we suggest interpreting these rankings as a screening analysis rather than as evidence that any individual feature is independently causal.

\begin{figure*}[htbp]
\centering{
\includegraphics[width=\textwidth]{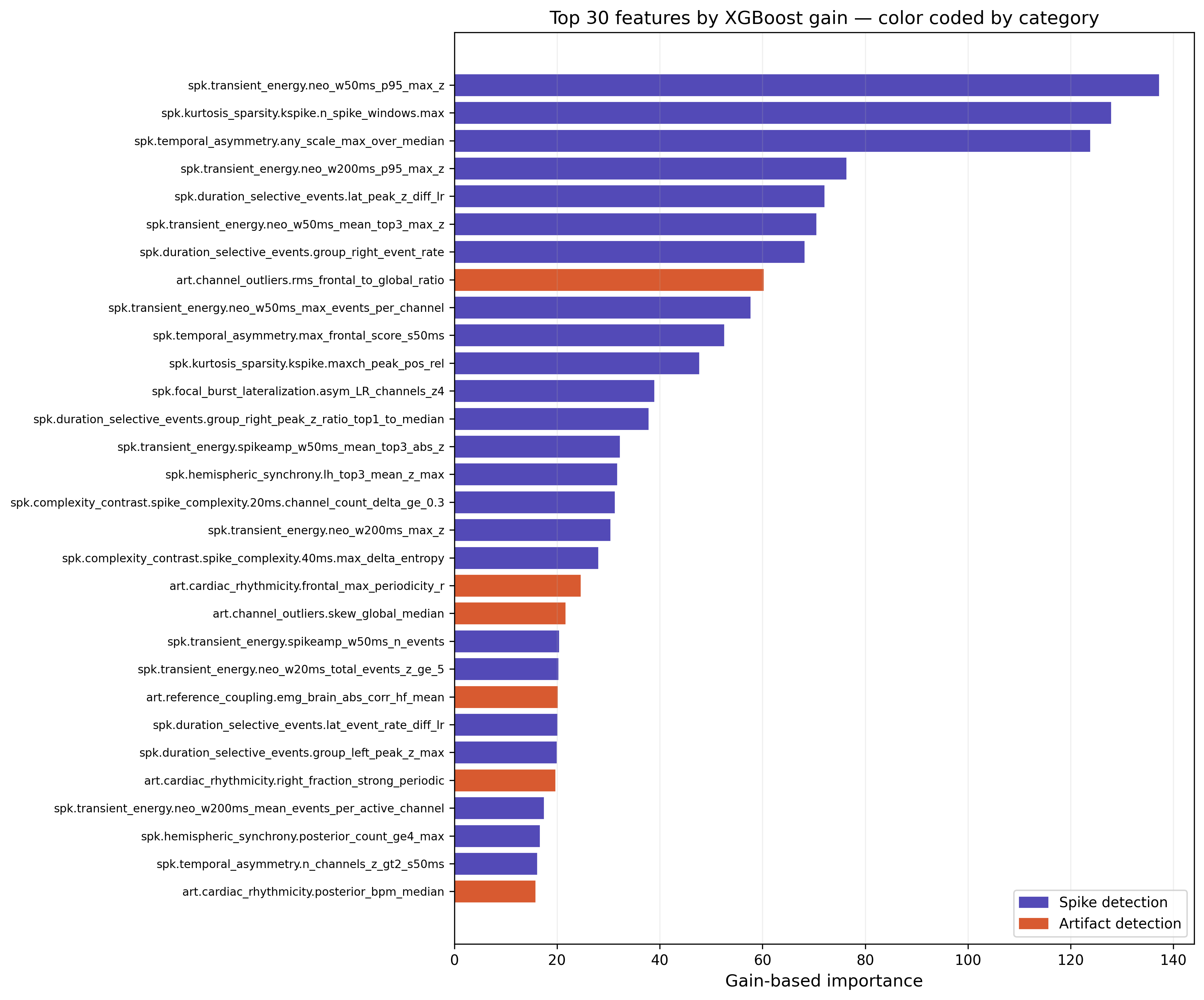}}
\caption{Top thirty features ranked by XGBoost gain-based importance and color coded by corresponding feature type (blue indicates spike-detection features, orange indicates artifact-detection features).}
\label{fig:feature_importance}
\end{figure*}

The top-ranked features spanned both spike-detection and artifact-detection families, indicating that the classifier did not rely on a single feature class. Morphologic and spatial spike features were prominent among the highest-importance variables, consistent with the clinical definition of IEDs as brief, sharply contoured events with characteristic field distributions. However, several artifact-aware features also appeared among the top thirty despite representing a minority of the full feature set. Specifically, despite representing only 18 percent of the total feature set, artifact-detection features such as frontal RMS ratio and cardiac rhythmicity metrics ranked highly in overall feature importance to the classifier (Fig.~\ref{fig:feature_importance}). This pattern supports the observation that artifact modeling improved classifier performance. 

These results also illustrate a practical advantage of the EEG-SpikeAgent workflow. Since each feature is implemented as a deterministic and named computation, high-importance variables can be traced back to specific signal operations, channel groups, and window-level summaries. This makes the trained classifier more inspectable than a purely latent representation and provides a starting point for future clinical review of individual decision rules, failure cases, and feature stability across datasets.

\section{Discussion}
\label{sec:discussion}
EEG-SpikeAgent demonstrates that LLM-guided program synthesis can produce an auditable EEG feature portfolio with strong threshold discrimination for IED detection.  The main contribution is an inspectable feature-search procedure and code representation, which is inherently more interpretable and traceable than the latent representations produced by pure deep-learning systems. Each candidate feature is implemented as deterministic code and can be directly inspected, ablated, and/or traced to model behavior. Artifact-aware features improved performance, suggesting that explicit artifact modeling is useful for agent-generated EEG feature engineering. 

The audit trail provides a practical advantage for model development. Each feature is implemented as executable code with a fixed configuration and run-level metrics. This makes the model easier to introspect relative to the latent representations that are typically produced by conventional deep learning methods, and allows performance changes to be traced to individual feature additions. Such traceability addresses a main limitation of automated deep learning EEG systems. Despite strong performance, deep learning models can be difficult to debug when they fail under new recording conditions, montage choices, or artifact distributions \cite{Janmohamed2022,Diniz2024Meta,Nhu2022Review}. In contrast, the features generated here retain clinically meaningful structure, including morphology, spatial aggregation, and artifact-coupling summaries, while still benefiting from automated exploration of a large design space.

The artifact-aware feature results suggest that artifact information can be useful beyond preprocessing.  Artifact features represented a minority of the final feature set, yet they improved balanced accuracy, F1 score, and sensitivity, and several ranked highly by gain-based importance. This result suggests that artifact modeling should not be treated only as a trivial preprocessing cleanup step. However, this 
interpretation remains hypothesis-generating. Gain-based importance is 
model-internal and individual artifact-aware features require review against EEG examples and error cases before being interpreted as clinically valid decision rules.

In relation to prior VEPISET results, EEG-SpikeAgent should be viewed as a complementary approach rather than as a direct replacement for optimized deep-learning architectures. Lin et al. reported at 80\% sensitivity a precision of 74.2\% and specificity of 97\% with a VGG network trained and evaluated on a single 80:20 training:test split of the VEPISET data \cite{Lin2025VEPISET}. They reported precision of 63.6\% and specificity of 95.0\% at 80\% sensitivity for the vEpiNet network, which was trained on an independent dataset. In comparison, EEG-SpikeAgent yielded mean precision of 47.0\% and mean specificity at 90.0\% at post-hoc operating point of 80\% sensitivity. We acknowledge that EEG-SpikeAgent did not match the operating-point performance of these deep-learning systems. However, EEG-SpikeAgent's contribution is that it provides an auditable feature-search procedure which is inherently more inspectable and traceable than pure deep-learning systems. Each candidate feature is implemented as deterministic code and can be inspected, ablated, and/or traced to model behavior.

The operating-point behavior of EEG-SpikeAgent is also informative for how agent-generated detectors might be used in practice. At the default threshold, the classifier produced very high specificity, indicating that the learned feature portfolio was conservative in labeling epochs as spike-containing. This behavior may be desirable in settings where false-positive burden is a major concern, such as retrospective review of large EEG corpora or triage systems intended to identify only definite events. However, clinical screening often requires higher sensitivity, even at the expense of additional false positives. The post-hoc threshold analysis therefore motivates and informs future work that incorporates calibration and threshold selection into detector design. In particular, operating points may be specified a priori according to the intended use case, such as high-specificity cohort enrichment or  high-sensitivity clinical screening, and could be paired with confidence-ranked event lists rather than binary labels alone.

We acknowledge several limitations of our findings. Importantly, while the system was highly specific, we acknowledge the limited sensitivity of 0.401 under the default classifier decision operating point (0.5). Future work will aim to improve calibration of the detector system. Secondly, the current analysis reports performance on curated 4 sec VEPISET epochs from a single public data source, with all IED subtypes collapsed into a binary Spike label. This setting does not test continuous-record screening, event localization, or performance under site-level distribution shift. In addition, the feature set is inspectable at the level of named computations, but inspectability does not guarantee interpretability or clinical validity. Individual high-importance features still require review against EEG examples and failure cases. Furthermore, statistics may vary substantially across acquisition systems, patient populations, and recording environments. Thus, future evaluations should test the stability of derived features. Lastly, our analysis did not stratify classifications by spike subtype, topology, and morphology. Future work should evaluate how performance is affected by these spike characteristics. Given the above limitations, we stress that the current study should therefore be viewed as proof of concept for agent-assisted EEG feature engineering on curated epochs, not as validation of a clinically ready spike detector.

EEG-SpikeAgent demonstrates a middle ground between manual symbolic pipeline design and opaque end-to-end learning. Generative models can propose candidate signal-processing code, while deterministic evaluation and auditable edit surfaces keep the resulting detector inspectable. This approach may extend naturally to other EEG tasks such as seizure detection, sleep staging, and artifact annotation.

\section*{Acknowledgment}
We thank the authors of VEPISET for making their dataset publicly available \cite{Lin2025VEPISET}.

\bibliographystyle{IEEEtran}
\bibliography{references}

\end{document}